# Medical Code Prediction from Discharge Summary: Document to Sequence BERT using Sequence Attention

Tak-Sung Heo*
*AI R&D Group*
*NHN Diquest*
Korea, Seoul
gjxkrtjd221@gmail.com

Yongmin Yoo*
*AI R&D Group*
*NHN Diquest*
Korea, Seoul
yooyongmin91@gmail.com

YeongJoon Park*
*AI R&D Group*
*NHN Diquest*
Korea, Seoul
yeongjoon1227@gmail.com

ByeongCheol Jo*
*AI R&D Group*
*NHN Diquest*
Korea, Seoul
byeongcheol7674@gmail.com

Kyounguk Lee
*AI R&D Group*
*NHN Diquest*
Korea, Seoul
arp1710@diquest.com

Kyungsun Kim
*AI R&D Group*
*NHN Diquest*
Korea, Seoul
kksun@diquest.com

*Equal contribution

*Abstract*— Clinical notes are unstructured text generated by clinicians during patient encounters. Clinical notes are usually accompanied by a set of metadata codes from the international classification of diseases (ICD). ICD code is an important code used in various operations, including insurance, reimbursement, medical diagnosis, etc. Therefore, it is important to classify ICD codes quickly and accurately. However, annotating these codes is costly and time-consuming. So we propose a model based on bidirectional encoder representations from transformers (BERT) using the sequence attention method for automatic ICD code assignment. We evaluate our approach on the medical information mart for intensive care III (MIMIC-III) benchmark dataset. Our model achieved performance of macro-averaged F1: 0.62898 and micro-averaged F1: 0.68555 and is performing better than a performance of the state-of-the-art model using the MIMIC-III dataset. The contribution of this study proposes a method of using BERT that can be applied to documents and a sequence attention method that can capture important sequence information appearing in documents.

I. INTRODUCTION

Clinical notes are unstructured text generated by clinicians during patient encounters. Clinical notes are usually accompanied by a set of metadata codes from the international classification of diseases (ICD), which present a standardized way of indicating diagnoses and procedures that were performed during the encounter with the patient. ICD is essentially a hierarchical classification that defines unique codes for patient conditions, diseases, infections, symptoms, causes of injury, and others. These unique diagnostic codes are assigned to patient records to facilitate clinical and financial decisions made by the hospital management for various tasks, including billing, insurance claims, and reimbursements [1, 2]. For these reasons, labeling the ICD code in the clinical description is very important. However, manually coding requires a lot of effort. It is not only error-prone but also time-consuming because it is what a person does. In addition, coding technical terms such as clinical records requires hiring people who can understand technical terms, which is costly to employ. In particular, the problem of multi-label classification, which is the task of automatically classifying multiple ICD codes in one clinical note, has received a lot of attention from the past to the present, and many studies have been conducted [3, 4, 5].

Recently, in natural language processing (NLP), many studies have been conducted using machine learning algorithms and deep learning algorithms to predict various classification tasks such as disease, prognosis, and ICD code using clinical notes [6, 7, 8, 9]. Kim et al. [6] used logistic regression, naïve Bayesian classification, single decision tree, and support vector machine to predict acute ischemic stroke using magnetic resonance imaging text reports of the brain. Heo et al. [7] used extreme gradient boosting, light gradient boosting machines, convolutional neural networks (CNN), and long short-term memory (LSTM) to predict stroke prognosis using magnetic resonance imaging text reports of the brain. Heo et al. [8] used CNN to predict the occurrence possibility of future atrial fibrillation using output texts of electrocardiogram extracted through an electrocardiography machine. Huang et al. [9] used feed-forward neural networks, CNN, LSTM, and gated recurrent units to predict ICD codes using clinical notes.

The medical information mart for intensive care III (MIMIC-III) is a large, single-center database comprising information relating to patients admitted to intensive care units (ICU) at a hospital and is a freely available benchmark dataset [10]. The dataset includes vital signs, medications, laboratory measurements, observations and notes charted by care providers, fluid balance, imaging reports, hospital length of stay, survival data, procedure and diagnoses codes providing ICD 9th revision (ICD-9) codes, and more. Some research papers have shown good performance by performing multi-label classification tasks using the MIMIC-III dataset [11, 12, 13].

In this study, we automatically assign ICD-9 codes through a deep learning model using clinical notes in the MIMIC-III data. Our model is based on bidirectional encoder representations from transformers (BERT) [14] which shows good performance in various tasks of NLP. We propose document-to-sequence BERT (D2SBERT), which solves the max sequence length problem that occurs when applying a transformer-based model to document classification tasks. And, we use sequence attention to capture important sequences for a specific ICD-9 code among many sequences appearing in the MIMIC-III dataset. As a result, our approach outperforms previous results on medical code prediction on the MIMIC-III dataset.

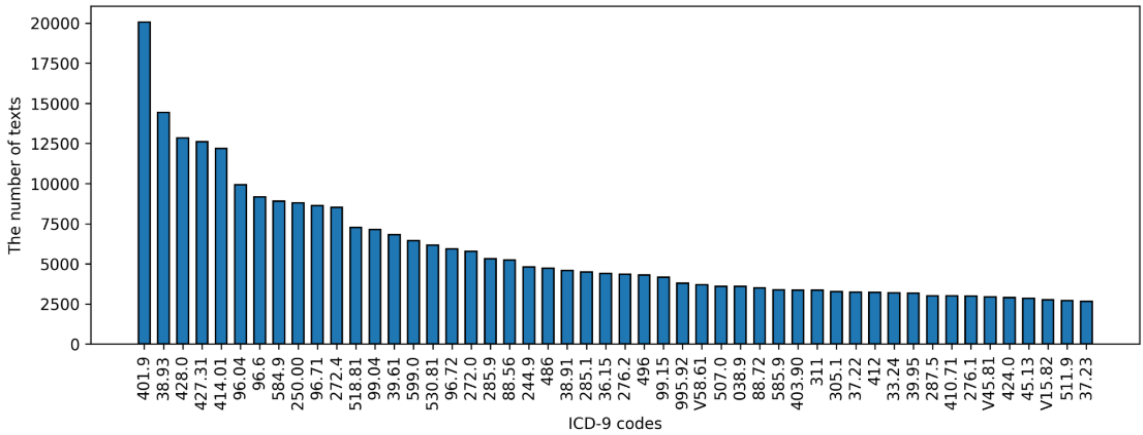

Fig. 1. Statistics of 50 most frequent ICD-9 codes

The contributions of this study are as follows:

- Since BERT has a maximum sequence length limitation when using pre-trained models, it is difficult to apply directly to documents having relatively long sequence lengths. To overcome these shortcomings, we introduce D2SBERT, which is BERT that can be applied to documents.

- We propose a sequence attention method that processes sequence representations extracted through D2SBERT. This method has the advantage of being able to capture important sequences among many sequences appearing in the document.

This study is composed as follows. Section 2 mentions related works about document classification. Section 3 describes the dataset and preprocessing method for the experiment. Section 4 explains the proposed methodology and details of the main components. Section 5 describes the hyperparameters used in the experiment and the experimental results. Finally, Section 6 mentions the summary of this research and discusses future studies.

## II. RELATED WORKS

Automatic ICD coding is a classification task for documents. Document classification tasks have been carried out in many studies and shown good performance using deep learning [11, 12, 13, 15].

Mullenbach et al. [11] proposed the convolutional attention for multi-label classification (CAML) and predicted several ICD-9 codes using 2,500 word tokens that appeared in the discharge summaries of the MIMIC-III dataset. Convolutional attention extracts feature maps, combined information of adjacent words, via CNN. It is also a neural network that predicts the target label using an important feature map corresponding to the target label among feature maps. Finally, they predicted ICD-9 codes by using sigmoid, an activation function, in the linear layer.

Hu and Teng [12] proposed the shallow and wide attention convolutional mechanism (SWAM), which broadened the network of CAML. SWAM is a neural network such as CAML, but it extracts more feature maps by increasing the number of filters on CNN used. They predicted multiple ICD-9 codes using 2,500 word tokens that appeared in the discharge summaries of the MIMIC-III dataset. Finally, they predicted ICD-9 codes by using sigmoid, an activation function, in the linear layer.

Mayya et al. [13] proposed the enhanced CAML (EnCAML), which consists of multi-channel. EnCAML is a method of utilizing CNNs with multi-Channel, which has the advantage of utilizing various information. They predicted multiple ICD-9 codes using 2,500 word tokens that appeared in the discharge summaries of the MIMIC-III dataset. Finally, they predicted ICD-9 codes by using an activation function named sigmoid in the linear layer.

Sun et al. [15] classified documents using BERT, which shows excellent performance in NLP. Since BERT has a max sequence length limitation, they truncated the text in three ways to fine-tune BERT with a maximum sequence length of 512 (510 sequence tokens and 2 special tokens) in the document. The first method is the BERT-head method, which uses 510 sequence tokens at the front, the second method is the BERT-tail method, which uses 510 sequence word tokens at the end, and the third method is the BERT-head-tail method, which uses 128 sequence tokens at the front and 382 sequence tokens at the end. In their experiments through BERT, the third method used both the front and the end parts of the document showed the best performance.

## III. DATASET

TABLE I. STATISTICS OF THE DISCHARGE SUMMARIES

| Description | Total |
|---|---|
| Discharge summaries | 52,726 |
| Words in the discharge summary | 79,801,402 |
| Average word count per discharge summary | 1513.51 |
| Unique words in the discharge summary | 150,853 |
| Words in the longest discharge summary | 10,500 |
| Words in the shortest discharge summary | 51 |
| ICD-9 codes (diagnoses and procedure) | 6,918 and 2,003 |

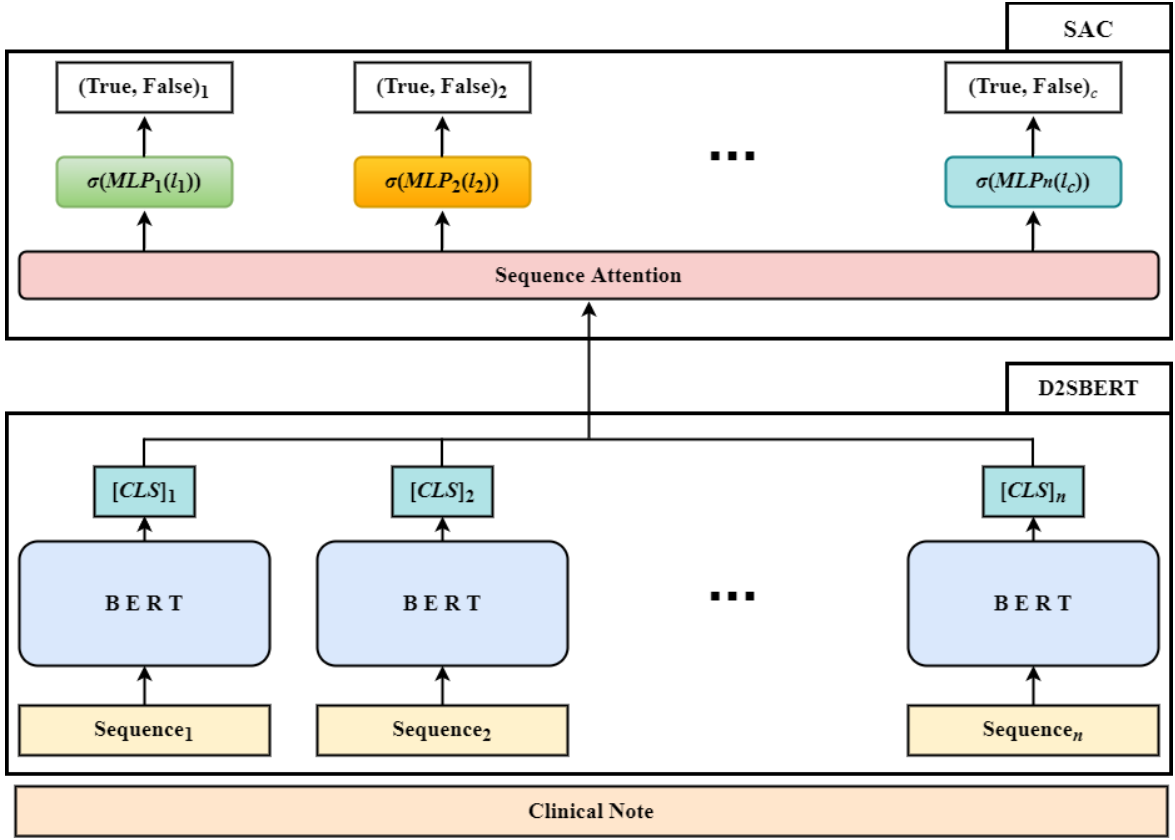

Fig. 2. Architecture of proposed model.

The MIMIC-III is a database comprising information relating to patients admitted to ICU at a hospital and consisted of 26 table data. Our study uses the *noteevents* table composed of text and the *diagnoses_icd* and *procedure_icd* tables containing ICD-9 code. The number of data used in this study is 52,726, as shown in Table I, and the experiment is conducted by dividing it into training set: validation set: test set = 8: 1: 1.

*A. Preprocessing*

The texts that appear in the *noteevents* table only use the discharge summaries corresponding to the first hospitalization, not all texts. And, all letters are changed to lowercase letters, and letters that did not consist of the alphabet are removed (e.g., removing '200' but keeping '200cc'). Although our proposed model can use all word tokens, we truncated the documents to a maximum length of 2,500 word tokens at the front part rather than using all tokens in the document, such as in previous studies [11, 12, 13]. In this way, we can ensure fairness in comparing the performance of the proposed model with previous studies.

The ICD-9 codes shown in the *diagnoses_icd* and *procedure_icd* tables are merged into one table and the 50 most frequent ICD-9 codes are used among the entire ICD-9 codes. Fig. 1 shows the distribution of the 50 most frequent ICD-9 codes.

## IV. METHODOLOGY

In this study, we propose D2SBERT, which solves the max sequence length problem that arises when applying a transformer-based model to document classification tasks. Also, we propose sequence attention that recognizes the relationship contents of the document. As shown in Fig. 2, the model is divided into D2SBERT, which extracts and processes all information in a sentence as much as possible, and sequence attention classifier (SAC), which predicts labels by reflecting only important information.

*A. Document-to-Sequence BERT (D2SBERT)*

In the case of the text classification task, BERT adds 2 special tokens, which are [CLS] token located at the beginning of a sentence and [SEP] token located at the end of a sentence to the existing text and uses them as input values [14]. Existing BERT has a limit on length because of their specified max sequence length [16, 17]. Generally, input sequences are truncated when fine-tuning BERT with long sequences. However, it results in missing data and is especially inappropriate for medical code predication, which requires a detailed understanding of the sequence.

D2SBERT first divides the entire document into sequences having a constant size of $n$ equal lengths to solve this problem. Each divided sequence is used as an input for the BERT model. To extract information about sequence, we use the [CLS] token extracted through BERT. We define the [CLS] token extracted through each sequence as $CLS_i$. When $CLS_i$ has the dimension of $\mathbb{R}^h$ and the number of sequences is $n$, the document representation vector ($D$) containing all sequence information has the dimension of $\mathbb{R}^{h \times n}$. The $D$ is expressed as equation (1).

$$D = [CLS_1; CLS_2; ...; CLS_n] \quad (1)$$

*B. Sequence Attention Classifier (SAC)*

Sequence attention decides important parts in the document by utilizing the relationship of the sequences. In the sequence attention classifier, the model extracts the important information from $D$ by using the sequence attention mechanism. When the total number of labels to be classified is $c$, $D$ and sequence attention matrix ($S$) having the dimension of $\mathbb{R}^{h \times c}$ are utilized to generate attention weights ($\alpha$). $\alpha$ is obtained as in equation (2).

$$\alpha = softmax(\tanh(D^T S)) \quad (2)$$

In equation (2), $\alpha$ is calculated in the form of a score using $softmax$. The label representation vectors ($L = [l_1, l_2, l_3, \ldots, l_c]$) are calculated using attention weight, which extracts the important information from a document. $l_i$ has the dimension of $\mathbb{R}^{h \times 1}$. Equation (3) shows the process of calculating $l_i$.

$$l_i = \sum_{j=1}^{n} \alpha_{ij} D_j \quad (3)$$

Information needed to determine whether the target label is true or false is contained in $l_i$. After each $l_i$ is connected to a different fully-connected layer ($MLP$), the final score of each target label is calculated using sigmoid function ($\sigma$). The final score about each target label is calculated as equation (4, 5).

$$MLP_i(x) = \sum_{j=1}^{h} W_j x_j + b \quad (4)$$

$$Score_i = \sigma(MLP_i(l_i)) \quad (5)$$

$Score_i$ has between 0 and 1. Finally, true or false of the label is determined by equation (6).

$$Predict_i^{ICD-9} = \begin{cases} True & if\ Score_i > 0.5 \\ False & otherwise \end{cases} \quad (6)$$

## V. EXPERIMENT

The performance of the proposed model in this study is compared with CAML, SWAM, EnCAML, BERT-head, BERT-tail, and BERT-head-tail models mentioned in related works. CAML, SWAM, and EnCAML, which are models that apply word embedding, used BioWordVec pre-trained with clinical documents and clinical notes [18]. And, BERT-head, BERT-tail, BERT-head-tail, and the proposed model used BioBERT pre-trained with clinical documents [19].

### A. Hyperparameter

TABLE II. HYPERPARAMETER USED IN THE EXPERIMENT

| Description | Range | Optimal Value(s) |
|---|---|---|
| The number of Sequences | {10; 25; 50; 100; 250} | 25 |
| Word per sequence | {250; 100; 50; 25; 10} | 100 |
| Optimizer | Adam | Adam |
| Learning rate | {3e-5; 1e-5; 5e-6} | 1e-5 |
| Exponential decay rates | $\beta_1$=0.9, $\beta_2$=0.999 | $\beta_1$=0.9, $\beta_2$=0.999 |
| Loss | Binary Cross Entropy | Binary Cross Entropy |

Table II shows the hyperparameters of BERT used in this experiment, and the hyperparameters used in this experiment are the same as the optimal value(s).

### B. Evaluation Metrics

We use two standard metrics to measure and compare the performance of our models: macro-averaged F1 and micro-averaged F1. The two standard metrics are calculated using precision and recall. And precision and recall are calculated using true-positive (TP), false-positive (FP), and false-negative (FN).

Macro-averaged F1 is a standard metric that calculates the metric independently for each class and then averages it. Macro-averaged F1 is calculated as the following equation (7, 8).

$$Precision_{macro} = \frac{1}{C} \sum_{c=1}^{C} \frac{TP_c}{TP_c + FP_c} \quad (7)$$

$$Recall_{macro} = \frac{1}{C} \sum_{c=1}^{C} \frac{TP_c}{TP_c + FN_c} \quad (8)$$

Micro-averaged F1 calculates an average metric by aggregating the contributions of all classes. Micro-averaged F1 is calculated as the following equation (9, 10).

$$Precision_{micro} = \frac{\sum_{c=1}^{C} TP_c}{\sum_{c=1}^{C} (TP_c + FP_c)} \quad (9)$$

$$Recall_{micro} = \frac{\sum_{c=1}^{C} TP_c}{\sum_{c=1}^{C} (TP_c + FN_c)} \quad (10)$$

Finally, macro-averaged F1 and micro-averaged F1 are calculated as the harmonic mean of precision and recall, such as equation (11).

$$F1_{score} = 2 \times \frac{Precision \times Recall}{Precision + Recall} \quad (11)$$

### C. Results

TABLE III. PERFORMANCE COMPARISON OF THE MODEL

| Model | F1 | |
|---|---|---|
| | *Macro* | *Micro* |
| CAML [10] | 0.56924 | 0.64993 |
| SWAM [11] | 0.58025 | 0.65994 |
| EnCAML [12] | 0.59653 0.6109* | 0.66594 0.6764* |
| BERT-head [14] | 0.49376 | 0.56627 |
| BERT-tail [14] | 0.45453 | 0.54011 |
| BERT-head-tail [14] | 0.49362 | 0.56566 |
| Proposed model | **0.62898** | **0.68555** |

Table III is the performance comparison of the model experimented under the same conditions except performance marked with an asterisk for 50 most frequent ICD-9 codes.

And, EnCAML marked with an asterisk (*) is the performance shown in the paper by Mayya et al. [13]. The experiment of Mayya et al. [13] marked with an asterisk corrected typos appearing in the MIMIC-III text data and also used the Fisher-Jenks Natural Breaks algorithm to further improve the performance of the EnCAML model. And, in Table III, the models labeled BERT use 510 sequence tokens and 2 special tokens. BERT-head uses 510 sequence tokens at the front part, BERT-tail uses 510 sequence tokens at the end part, and BERT-head-tail uses 128 sequence tokens at the front part and 382 sequence tokens at the end part.

For BERT, the BERT-head method showed the highest performance. This indicates that there is more important content about ICD-9 codes in the front part than the end part of the document. However, BERT, which generally shows good performance in NLP, showed significantly lower performance than other models. This is because it has the max sequence length limit of BERT, so the method using BERT cannot fully understand all contents of the document because it uses only a part of the document.

In the model using CNN-based attention, EnCAML showed high performance compared to CAML and SWAM models because it uses various semantic information.

It can be seen that the proposed model in this study has the higher performance of macro-averaged F1: 0.13522 to 0.17445 and micro-averaged F1: 0.11928 to 0.14544 compared to BERT. This result shows that we can overcome the disadvantage of BERT with max sequence length limitation. And, we can see that the proposed model has a higher performance of macro-averaged F1: 0.03245 to 0.05974 and micro-averaged F1: 0.01961 to 0.03562 compared to the models using CNN-based attention. Moreover, although we do not use additional typo correction and other algorithms, the proposed model outperforms EnCAML, the state-of-the-art model, that uses the typo correction and the Fisher-Jenks Natural Breaks algorithm.

## VI. CONCLUSION

ICD code is an important code used in a variety of operations, including insurance, reimbursement, medical diagnosis, etc. Therefore, it is important to classify ICD codes from clinical notes quickly and accurately. However, manual coding is time-consuming and costly. Therefore, we propose a deep learning model that automatically assigns ICD-9 codes using clinical notes from MIMIC-III data in this research. Because BERT has max sequence length limitations, we used D2SBERT, which divides the document into sequences and applies BERT to each sequence. Also, sequence attention is applied to features with sequence information extracted via D2SBERT, and finally, ICD-9 code is predicted through a fully connected layer. Experiments show that the proposed model has the highest performance compared to the previous research models.

We have achieved useful results in the clinical field using clinical notes called MIMIC-III. Although this study aims at the clinical field, our method can be applied to various multi-label classification fields, such as the cooperative patent classification, the international patent classification for management strategy, and sentiment analysis to predict complex sentiments. Based on the experimental results, we expect that it will show good performance on other datasets.

Since the clinical note of MIMIC-III is a document written by several clinicians, the sentences have a structure that is not segmented correctly. So, we found a proper sequence length per document throughout the experiment. However, obtaining the proper sequence length is time-consuming because it is obtained through many experiments. Therefore, if we find a more efficient way to split the sentence, not the sequence split way, we can save time, and the performance of the model is expected to increase. Thus, in the future study, we will explore proper sentence segmentation methods and use various language models that outperform BERT to improve performance more.

ACKNOWLEDGMENT

This work was supported by the ICT R&D By the Institute for Information & communications Technology Promotion(IITP) grant funded by the Korea government(MSIT) [Project Number : 2020-0-00113, Project Name : Development of data augmentation technology by using heterogeneous information and data fusions] and the Industrial Technology Innovation Program funded by the ministry of Trade, Industry & Energy(MOTIE, Korea) [Project Number : 20008625, Project Name : Development of deep tagging and 2D virtual try on for fashion online channels to provide mixed reality visualized service based on fashion attributes]

REFERENCES

[1] P. B. Jensen, L. J. Jensen, and S. Brunak, "Mining electronic health records: towards better research applications and clinical care". *Nat. Rev. Genet.*, 13(6):395-405.

[2] M. Li, Z. Fei, M. Zeng, F. X. Wu, Y. Li, Y. Pan, and J. Wang, "Automated ICD-9 coding via a deep learning approach". *IEEE/ACM Trans. Comput. Biol Bioinform.*, 16(4):1193-1202.

[3] A. Avati, K. Jung, S. Harman, L. Downing, A. Ng, and N. Shah, "Improving palliative care with deep learning". *BMC Med. Inform. Decis.*, 18(4):55-64.

[4] E. Birman-Deych, A. D. Waterman, Y. Yan, D. S. Nilasena, M. J. Radford, and B. F. Gage, "Accuracy of ICD-9-CM codes for identifying cardiovascular and stroke risk factors". *Med. Care.*, 480-485.

[5] D. Zhang, D. He, S. Zhao, and L. Li, "Enhancing automatic icd-9-cm code assignment for medical texts with pubmed". In *BioNLP*. Canada, pp. 263-271, August 2017.

[6] C. Kim, V. Zhu, J. Obeid, and L. Lenert, "Natural language processing and machine learning algorithm to identify brain MRI reports with acute ischemic stroke". *PLoS One.*, 14(2):e0212778.

[7] T. S. Heo, C. Kim, J. M. Choi, Y. S. Jeong, and Y. S. Kim, "Various Approaches for Predicting Stroke Prognosis using Magnetic Resonance Imaging Text Records". In *Proc. 3rd Clinical NLP.*, pp. 1-6, November 2020.

[8] T. S. Heo, C. Kim, J. D. Kim, C. Y. Park, and Y. S. Kim, "Prediction of Atrial Fibrillation Cases: Convolutional Neural Networks Using the Output Texts of Electrocardiography". *Sens. Mater.*, 33(1):393-404.

[9] J. Huang, C. Osorio, and L. W. Sy, "An empirical evaluation of deep learning for ICD-9 code assignment using MIMIC-III clinical notes". *Comput. Meth. Prog. Bio.*, 177:141-153.

[10] A. E. Johnson, T. J. Pollard, L. Shen, H. L. Li-wei, M. Feng, M. Ghassemi, B. Moody, P. Szolovits, L. A. Celi, and R. G. Mark, "MIMIC-III, a freely accessible critical care database". *Sci. Data.*, 3(1):1-9.

[11] J. Mullenbach, S. Wiegreffe, J. Duke, J. Sun, and J. Eisenstein, "Explainable Prediction of Medical Codes from Clinical Text". In *Proc. HLT-NAACL*. United States, vol. 1, pp. 1101-1111, June 2018.

[12] S. Y. Hu, and F. Teng, "An Explainable CNN Approach for Medical Codes Prediction from Clinical Text". *arXiv preprint.*, arXiv:2101.11430.

[13] V. Mayya, S. Kamath, G. S. Krishnan, and T. Gangavarapu, "Multi-channel, convolutional attention based neural model for automated

diagnostic coding of unstructured patient discharge summaries". *Future Gener. Comput. Syst.*, 118:374-391.

[14] J. Devlin, M. W. Chang, K. Lee, and K. Toutanova, "BERT: Pre-training of Deep Bidirectional Transformers for Language Understanding". In *Proc. HLT-NAACL.* United States, vol. 1, pp. 4171-4186, June 2019.

[15] C. Sun, X. Qiu, Y. Xu, and X. Huang, "How to fine-tune BERT for text classification?". In *CCL.* China, pp. 194-206, October 2019.

[16] Z. Dai, Z. Yang, Y. Yang, J. G. Carbonell, Q. Le, and R. Salakhutdinov, "Transformer-XL: Attentive Language Models beyond a Fixed-Length Context". In *Proc. ACL.* Italy, pp. 2978-2988, July 2019.

[17] I. Beltagy, M. E. Peters, and A. Cohan, "Longformer: The long-document transformer". *arXiv preprint*. arXiv:2004.05150.

[18] Y. Zhang, Q. Chen, Z. Yang, H. Lin, and Z. Lu, "BioWordVec, improving biomedical word embeddings with subword information and MeSH". *Sci. Data.*, 6(1):1-9.

[19] J. Lee, W. Yoon, S. Kim, D. Kim, S. Kim, C. H. So, and J. Kang, "BioBERT: a pre-trained biomedical language representation model for biomedical text mining". *Bioinformatics.*, 36(4):1234-1240.